\documentclass[11pt]{article}

\usepackage[]{acl}

\usepackage{times}
\usepackage{latexsym}

\usepackage[T1]{fontenc}
\usepackage[utf8]{inputenc}

\usepackage{microtype}

\usepackage{inconsolata}

\usepackage{graphicx}

\usepackage{xcolor}
\usepackage{verbatim}
\usepackage{graphicx}
\usepackage{url} % typeset URL's reasonably
\usepackage{listings}
\usepackage{amsmath} 
\usepackage{hyperref}
\usepackage{array}
\usepackage{booktabs} % Add this to your preamble if not already included
\usepackage{array}    % Add this to your preamble for m{} vertical centering
\usepackage[most]{tcolorbox}
\definecolor{verylightgray}{gray}{0.97} % 0 = black, 1 = white. 0.97 = very light gray.
\usepackage{amsmath}
\usepackage{float}

\usepackage{fontspec}
\usepackage[english,bidi=default]{babel}

\babelfont{rm}[
  BoldFont = texgyretermes-bold.otf,
  ItalicFont = texgyretermes-italic.otf,
  BoldItalicFont = texgyretermes-bolditalic.otf
]{texgyretermes-regular.otf}

\babelprovide[import]{arabic}
\babelfont[*arabic]{rm}[
  Scale=0.80,
  BoldFont = Amiri-Bold.ttf,
  ItalicFont = Amiri-Italic.ttf,
  BoldItalicFont = Amiri-BoldItalic.ttf
]{Amiri-Regular.ttf}

\renewcommand{\arraystretch}{1.2} % this is for the tables

\title{Ara-HOPE: Human-Centric Post-Editing Evaluation for Dialectal Arabic to Modern Standard Arabic Translation}

\author{
 \textbf{Abdullah Alabdullah\thanks{The previous affiliation for Abdullah Alabdullah (where most of this research was completed) is the University of Manchester, Manchester, M13 9PL, The United Kingdom.}\textsuperscript{1}},
 \textbf{Lifeng Han\textsuperscript{2}},
 \textbf{Chenghua Lin\textsuperscript{3}},
\\
\\
 \textsuperscript{1}School of Informatics, University of Edinburgh, The United Kingdom\\
 \textsuperscript{2}Leids Universitair Medisch Centrum \& LIACS, Universiteit Leiden, NL\\
 \textsuperscript{3}School of Computer Science, University of Manchester, The United Kingdom
\\
 \small{
   \textbf{Correspondence:} \href{mailto:a.alabdullah@sms.ed.ac.uk}{a.alabdullah@sms.ed.ac.uk} \& \href{mailto:l.han@lumc.nl}{l.han@lumc.nl,liacs.leidenuniv.nl }
 }
}
\begin{document}
\maketitle
\begin{abstract}

% one short sentence for motivation and setting the scene
% what we did overall (overall achievement)
% What did we do 1 (first achievement - corresponding to TASK1) + briefly How did we do it + What did we find
% What did we do 2 (first achievement - corresponding to TASK1) + briefly How did we do it + What did we find
% What do we think it means (significance)

Dialectal Arabic to Modern Standard Arabic (DA-MSA) translation is a challenging task in Machine Translation (MT) due to significant lexical, syntactic, and semantic divergences between Arabic dialects and MSA. Existing automatic evaluation metrics and general-purpose human evaluation frameworks struggle to capture \textbf{dialect-specific MT errors}, hindering progress in translation assessment. This paper introduces \textit{Ara-HOPE}, a human-centric post-editing evaluation framework designed to systematically address these challenges. The framework includes a five-category error taxonomy and a decision-tree annotation protocol. Through comparative evaluation of three MT systems (Arabic-centric Jais, general-purpose GPT-3.5, and baseline NLLB-200), Ara-HOPE effectively highlights systematic performance differences between these systems. Our results show that dialect-specific terminology and semantic preservation remain the most persistent challenges in DA-MSA translation. Ara-HOPE establishes a new framework for evaluating Dialectal Arabic MT quality and provides actionable guidance for improving dialect-aware MT systems. For reproducibility, we make the annotation files and related materials publicly available at \url{https://github.com/abdullahalabdullah/Ara-HOPE}

% .
% Dec 19 EACL-WS-dia \url{https://sites.google.com/view/vardial-2026/call-for-papers} 
% thesis \url{https://www.overleaf.com/project/680f5823c2e3f20a2de03e4b} This document is a supplement to the general instructions for *ACL authors. It contains instructions for using the \LaTeX{} style files for ACL conferences.
% The document itself conforms to its own specifications, and is therefore an example of what your manuscript should look like.
% These instructions should be used both for papers submitted for review and for final versions of accepted papers.
\end{abstract}

\section{Introduction}
This paper enhances Dialectal Arabic to Modern Standard Arabic (DA-MSA) translation quality assessment through a human-centric evaluation framework, addressing key gaps in current methods for under-resourced language pairs. %Building on paper 2’s architectural analysis, It introduces and validates an error taxonomy and annotation protocol designed to capture \textit{dialect-specific translation failures} that automated evaluation metrics often miss.

Dialectal Arabic (DA) refers to the informal varieties of Arabic used in everyday communication, which vary significantly across regions. In contrast, Modern Standard Arabic (MSA) is the formal variety used in writing, education, and traditional media \cite{diab2010colaba}.

Unlike interlingual translation tasks (e.g. translating from English to Arabic), DA-to-MSA translation is a dialect normalization task that introduces additional challenges arising from divergences between Arabic dialects and MSA. Key differences between DA and MSA that affect MT include: (i) \textbf{Orthographic differences}: Dialects do not follow standardized spelling rules and the same word may appear in different forms, making automatic text normalization difficult for NLP systems \cite{alhafni-etal-2024-exploiting}.
(ii) \textbf{Morphological differences}: Morphology is how words change form to express features such as tense or gender. While MSA has a rich, complex, and standardized morphology, spoken dialects often simplify these systems by omitting rules or using reduced forms \cite{KIRCHHOFF2006589}.
(iii) \textbf{Lexical differences}: Dialectal vocabulary often includes slang and idiomatic expressions that are typically absent in MSA \cite{hadj-mohamed-etal-2023-alphamwe}.
(iv) \textbf{Syntactic differences}: Syntax, sentence structure, and word order in many dialects differ from MSA \cite{10.5555/1621774.1621784}.
(v) \textbf{Code-switch}: Speakers often mix DA and MSA, and sometimes even foreign words, within a single sentence. This adds more complexity when trying to build systems that automatically translate from DA to MSA \cite{hamed-etal-2025-survey}. These challenges make DA-MSA translation a complex task requiring specialized approaches grounded in a sound understanding of common error types in DA-MSA translation systems.

In this paper we introduce a post-editing human evaluation framework that offers the granularity needed to identify systemic weaknesses in DA-MSA machine translation systems. Unlike general-purpose frameworks, our proposed framework (Ara-HOPE) targets translation errors that result from DA-specific translation challenges.

% {Our Contributions}:
% \begin{enumerate}
%     \item \textbf{Ara-HOPE Taxonomy}: A five-category error classification targeting DA-specific challenges, such as colloquial expression handling and target-language adaptation, validated on 200 annotated examples from the Dial2MSA-Verified corpus.
%     \item \textbf{Cognitive Workflow Optimization}: A decision-tree annotation protocol designed to reduce cognitive load for non-expert evaluators through hierarchical error categorization.
%     \item \textbf{Architectural Diagnostics}: A comparative framework quantifying performance differences across Arabic-centric (Jais), general-purpose (GPT-3.5), and multilingual baseline (NLLB-200) models using severity-weighted error profiles.
% \end{enumerate}

\section{Related Work}
\subsection{Advancements in Dialectal Arabic Translation}
Neural architectures have transformed DA-MSA translation by capturing contextual dependencies and complex syntactic and semantic relationships, effectively modeling dialectal variations and producing more fluent and accurate translations \cite{9173dc9888094080b48ab529a68bbc63}. The emergence of large parallel corpora has further advanced Neural Machine Translation by mitigating data scarcity and enabling training on dialectally diverse data. A key example is the MADAR corpus, which contains parallel translations from 25 Arabic city dialects \cite{Bouamor2018TheMA}. Recent studies have compared neural architectures, including encoder–decoder models like NLLB-200 and decoder-only models like GPT-3 and GPT-4o \cite{nllbteam2022languageleftbehindscaling,NEURIPS2020_1457c0d6,alabdullah2025advancingdialectalarabicmodern}. Decoder-only models have demonstrated better performance at preserving cultural context \cite{yakhni-chehab-2025-llms}.

The emergence of LLMs trained on large and diverse multilingual data, and further optimized through instruction tuning, has enabled models to follow natural-language instructions and generate appropriate outputs for a wide range of tasks, including machine translation, without costly task-specific fine-tuning \cite{NEURIPS2020_1457c0d6}. This paradigm is known as In-Context Learning (ICL), where the desired task is specified directly in the prompt, and the model infers the mapping from the provided context. In zero-shot ICL, the model is instructed to translate from a source to a target language without providing any in-prompt examples. Zero-shot prompting has been particularly effective when parallel data is scarce and is widely used in DA-MSA shared tasks such as OSACT \cite{osact2024} and NADI \cite{abdulmageed2024nadi}.

While multilingual LLMs capture general linguistic features through large-scale multilingual pretraining, new Arabic-specialized LLMs like Jais improved the handling of dialectal nuances and cultural references \cite{sengupta2023jaisjaischatarabiccentricfoundation, mousi-etal-2025-aradice}, producing more natural and contextually appropriate translations. Despite these advances, neural models continue to underperform on DA translation due to persistent challenges with dialectical nuances and culturally embedded expressions \cite{mousi-etal-2025-aradice,alabdullah2025advancingdialectalarabicmodern}.

\subsection{Dialectal Arabic Translation Evaluation}

% auto evaluation metrics that exisit and their limittaions

% human evaluation frameworks and their limitations + why they do not work for DA

% the nees for DA-spealized human evaluation framwrok

% \subsubsection{Automatic Evaluation Metrics and Their Limitations}

Traditional automatic evaluation metrics such as BLEU \cite{post-2018-call} and METEOR \cite{banerjee-lavie-2005-meteor} are limited for DA-MSA translation, as they rely on lexical overlap and perform poorly on morphologically rich languages like Arabic. \newcite{Bouamor2014AHJ} proposed AL-BLEU, an extension of BLEU that assigns partial credit for stem and morphological matches, yielding better correlation with human judgment than standard metrics. However, AL-BLEU remains a lexical-overlap metric and fails to capture semantic adequacy, particularly an issue for syntactically flexible languages like Arabic. While metrics like BLEU allow evaluation against multiple reference translations, these are challenging to produce. As a result, these metrics favor literal translations with high lexical overlap over contextually appropriate ones that better reflect human judgment, but differ lexically from the reference.

In low-resource settings such as dialectal Arabic, neural evaluation metrics like BERTScore \cite{zhang2019bertscore} and COMET \cite{rei-etal-2020-comet} also face challenges. \newcite{falcao2024comet} showed that COMET’s performance on under-resourced languages is constrained by imbalanced training data. These metrics rely on pretrained models which typically lack sufficient dialectal Arabic training data, leading to lower-quality embeddings.

Human evaluation frameworks such as the Multidimensional Quality Metrics (MQM) assess fine-grained translation errors like omissions and register mismatches \cite{lommel2014mqm,lommel-etal-2024-multi}. While MQM allows for more precise diagnostic error analysis due to its detailed error taxonomy, this comes at the cost of increased complexity in the annotation framework, requiring extensive annotator training, which can be costly and difficult to achieve for low-resource languages \cite{kocmi-etal-2024-error}. Moreover, DA translation requires a specialized human evaluation framework that captures the most impactful error types and minimizes subjectivity in assessing translation quality for this specific task, with only minimally trained annotators.
% To further advance the research in this direction, we will build our methodology on top of the HOPE metric \cite{gladkoff2022hope}, a human-centric post-editing metric simplified from MQM, which has eight error types and error severity-level design.
To further advance research in this direction, we build our methodology on the HOPE metric \cite{gladkoff2022hope}. HOPE is a task-oriented evaluation framework designed to address the limitations of both automatic evaluation metrics (e.g. BLEU) and highly fine-grained but complex human evaluation frameworks such as MQM. Ara-HOPE is human-centric because it only incorporates eight evaluation criteria that capture the most critical and recurring errors in translation between the Levantine Arabic dialect and MSA, reflecting translation quality as perceived by native speakers. The assigned error scores correspond to the post-editing effort required to bring the translation to an acceptable quality.

\section{Methodology Design}
\subsection{Framework Development}

\textbf{Ara-HOPE Error Taxonomy Design:}
Developing a robust human evaluation framework requires clearly defined objectives. For DA-MSA translation, this requires an error taxonomy tailored to the specific challenges of this language pair. While the HOPE framework \cite{gladkoff2022hope} offers a foundation for general post-editing translation assessment, we adapt the general error types in HOPE to capture the specific challenges of DA-MSA translation. To minimize subjective judgment between annotators, we also reduce the severity scoring range to three levels: from 0 (no errors) to 2 (major error). Our proposed taxonomy is designed to evaluate translation quality, identify system weaknesses, and guide system improvements, while remaining usable by native DA speakers without requiring extensive annotation training. To achieve this, we employ a direct quality estimation approach that evaluates predefined aspects of translation quality (e.g. fluency) at the segment level using discrete severity scores. This design makes our framework easier to train annotators on and faster to apply than MQM.

\textbf{Design Principles:}
To ensure a theoretically sound and practical taxonomy, we developed guidelines grounded in best practices in translation quality assessment \cite{Han02072020,10.1007/s10579-021-09537-5} and refined them through iterative feedback from focus groups with native DA speakers. The finalized guidelines comprise six core principles: (1) \textbf{Identifiability:} Errors must be detectable by minimally trained native dialect speakers to ensure consistent evaluation. 
(2) \textbf{Distinguishability:} Categories should have minimal conceptual overlap, with clear definitions to avoid confusion.
(3) \textbf{Actionability:} Error classification must enable targeted system improvements through quantifiable error severity and clear mapping to issues in the translation system.
(4) \textbf{Comprehensiveness:} The taxonomy must capture all major errors common in DA-MSA MT, including dialect-specific challenges.
(5) \textbf{Relevance:} Categories must address issues unique to DA-MSA translation rather than pure general translation errors.
(6) \textbf{Usability:} The taxonomy should remain manageable in size, balancing high-level categories with sufficient granularity.

% These principles are summarized in Table~\ref{tab:taxonomy_principles}.

\textbf{Taxonomy Structure:}
The Ara-HOPE taxonomy categorizes DA-MSA translation errors into three hierarchically structured classes:
(1) \textbf{Fluency Error (FLU):} Grammatical or linguistic errors in the MSA translation, independent of the DA source sentence. 
(2) \textbf{Meaning Transfer Error:} Failures to accurately preserve source meaning. This category includes Proper Name (PRN) errors, referring to incorrect translations of names of people, places, or organizations; Dialect-Specific Term (TRM) errors, which involve untranslated or mistranslated dialectal expressions that alter meaning; and General Semantic Mistranslation (GSMIS) errors, which includes omissions, additions, or other semantic changes.
(3) \textbf{Adaptation Error (ADP):} Translations that are unnatural or contextually inappropriate in tone, style, or intent.

GSMIS captures global meaning changes arising from the model's inability to handle DA’s contextual dependencies and includes all meaning changes not covered by PRN or TRM. This category was introduced after observing that LLMs often produce semantic distortions beyond proper name or dialect-specific term mistranslations.

\textbf{Decision Tree Implementation}
While a list format effectively outlined the taxonomy, it was later reformatted as a decision tree to reduce cognitive load for our minimally trained evaluators. The decision tree guides annotators through hierarchical error categories, as shown in Figure~\ref{fig:arabic-annotation-tree} in the Appendix. An English translated version of the tree is provided in Figure~\ref{fig:annotation-tree} in the Appendix.

This structure simplifies annotation by providing step-by-step guidance, helping evaluators understand the evaluation process, handle multiple error types systematically, and make consistent judgment at each level. The tree begins with the three primary categories (Fluency, Meaning Transfer, and Adaptation) and expands into more granular subcategories. Each node presents a yes/no question to minimize subjective judgments. Since some error types (e.g. Dialect-Specific Term) can be challenging to distinguish from other categories, we provided annotators with a table of annotation guidelines with illustrative examples to clarify the distinctions between error types. The guidelines can be found in Figure~\ref{fig:annotation-guidelines} in the Appendix.

Practical initial testing demonstrated Ara-HOPE’s effectiveness in addressing dialect-specific translation challenges and improving annotation consistency. The decision-tree structure also proved especially effective in guiding evaluators and enhancing usability.

%===================
\subsection{Dataset  Preparation}

Our human evaluation experiment utilized 200 tweets from the Levantine development set of the Dial2MSA-Verified dataset \cite{khered2025dial2msa}, a high-quality parallel corpus for DA-MSA translation in the social media domain. It extends the original Dial2MSA dataset \cite{mubarak2018dial2msa} and applies automated corrections and human evaluation by native speakers to produce reliable MSA references.

This tweets dataset was chosen for its comprehensive representation of DA-MSA translation challenges, including:
\textbf{Lexical variations:} colloquial and multi-word expressions.
\textbf{Semantic shifts:} cultural references and context-dependent meaning.
\textbf{Orthographic variations:} common in Levantine social media text.

The tweets in this dataset span a wide range of tones, from casual tweets to heated discourse, ensuring thorough testing of MT systems' ability to handle DA translation, while preserving sentiment. Moreover, the dataset allows for testing Ara-HOPE's capacity to capture nuanced dialect-specific translation errors.

Our 200-example subset size exceeds the lower bound below which translation quality estimates become unreliable, addressing concerns that a very small sample may lead to unreliable error analysis \cite{gladkoff2022uncertainty}.

\subsection{Translation Systems Selection}
For this human evaluation experiment, we selected three translation systems using three predefined criteria, which ensure a robust comparative analysis of DA-MSA translation performance:
(i) \textbf{Reliability and Proven Performance}: Each system has demonstrated strong performance on DA-MSA translation in prior work.
(ii) \textbf{Architectural Diversity}: The systems represent distinct model families (encoder-decoder vs.\ decoder-only) and differ in pretraining data sources, encouraging diverse outputs and reducing redundancy.
(iii) \textbf{Performance Scope}: We include both advanced and baseline systems to contrast general-purpose and Arabic-specialized approaches.

{The Selected Systems Are:}
\textbf{Jais}: A state-of-the-art Arabic-centric model trained with approximately one-third of its training tokens drawn from Arabic data, excelling at capturing DA-MSA nuances \cite{sengupta2023jaisjaischatarabiccentricfoundation}.
\textbf{GPT-3.5}: A general-purpose multilingual LLM. We will use this model to benchmark multilingual systems against an Arabic-specialized model. This model was introduced as an improvement over its predecessor, GPT-3 \cite{NEURIPS2020_1457c0d6}.
\textbf{NLLB-200 3.3B}: A multilingual baseline MT system, providing contrast and revealing errors that more advanced systems may avoid \cite{nllbteam2022languageleftbehindscaling}.

Each of the three systems was prompted to translate the same 200 examples. A zero-shot instruction setup was used to fairly evaluate each model’s baseline capabilities without fine-tuning or advanced prompting configuration. Jais and GPT-3.5 received consistent prompts (in Arabic for Jais and English for GPT-3.5) while NLLB-200 required no prompting due to its built-in MT task support.

\section{Implementation}

\subsection{Annotator Selection and Training}
Two native speakers of Syrian Levantine Arabic with advanced proficiency in MSA volunteered as annotators. Both hold undergraduate degrees in Arabic Language, enabling them to identify violations of MSA grammatical rules. Recruiting two annotators allowed for inter-annotator agreement (IAA) analysis while keeping the annotators group size manageable.

Annotators received initial training through a 25-minute pre-recorded video and supplementary materials. Training covered: (1) an \textbf{Error Taxonomy Guide}: providing detailed definitions of error types, alongside diverse examples of correct and incorrect DA-MSA translations, and (2) a \textbf{Decision Tree Workflow}: annotators first read the source and gold translation to establish the intended meaning, then highlighted errors by severity starting with the most critical, and finally used the decision tree (Figure~\ref{fig:annotation-tree} in the Appendix) to classify errors systematically.

Offline support was available throughout the annotation period. Each annotator spent approximately 12 hours on their tasks, supported by multiple feedback sessions to ensure clarity and consistency.

% \begin{itemize}
%     \item \textbf{Error Taxonomy Guide}: Detailed descriptions of error types (fluency, meaning transfer, adaptation) with diverse examples of correct and incorrect DA-MSA translations.
%     \item \textbf{Decision Tree Workflow}: Annotators were instructed to:
%     \begin{itemize}
%         \item Read the source and gold translation to understand the intended meaning before evaluating the machine translation.
%         \item Highlight errors by severity, beginning with the most significant.
%         \item Use the decision tree (see Figure~\ref{fig:annotation-tree}) to classify errors systematically and minimize bias.
%     \end{itemize}
% \end{itemize}

\subsection{Pilot Testing and Framework Refinement}

Before full-scale evaluation, a pilot study was conducted with one annotator annotating 10 DA-MSA examples. The primary goal was to test the framework’s usability and effectiveness. The feedback received focused on:
(1) \textbf{Instruction Clarity}: The annotator initially struggled to distinguish between meaning transfer and adaptation errors. The definitions were revised accordingly.
(2) \textbf{Questionnaire Usability}: The layout was changed to ensure ease of use and reduce cognitive load.
(3) \textbf{Severity Scale Adjustments}: Confusing mid-range scores were removed, leading to refinement of the scoring scale.

The pilot test confirmed the framework's overall effectiveness and identified minor improvements needed to enhance clarity and usability. Revisions ensured annotators could complete evaluation efficiently without sacrificing accuracy.

\subsection{Questionnaire Design}

\begin{figure*}[t]
\begin{center}
\includegraphics[width=\textwidth]{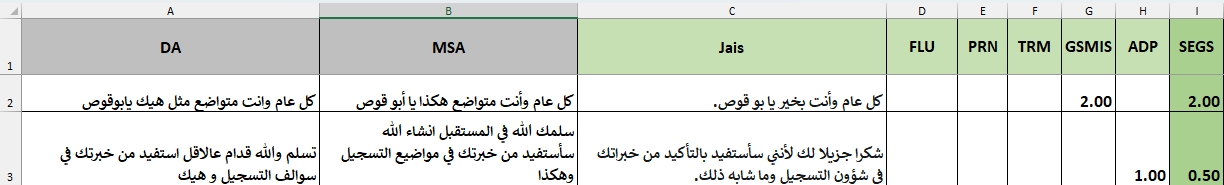} % Rotate by 90 degrees and scale to full page width
\end{center}
\caption[Annotation Questionnaire Design]
{Example layout of the annotation questionnaire used for evaluating DA-MSA translations across fluency, meaning transfer, and adaptation error categories. To avoid unfairly penalizing systems that preserved meaning but made adaptation errors, adaptation scores were weighted half in the segment total error score calculation.}
\label{fig:annotation-examples}
\end{figure*}

A structured Excel sheet was created to support annotation and data analysis. Each sheet contained three sets of columns, one per translation engine. Each annotator evaluated 600 translations in total (200 for each translation engine). Figure~\ref{fig:annotation-examples} shows two annotated examples for the Jais system. The first three columns show the DA sentence, the MSA gold translation, and the proposed machine translation. Columns 4-8 represent the five Ara-HOPE error categories, with annotators assigning severity scores from 0 (no error) to 2 (major error). Empty cells indicate a score of 0, and the final column sums the error scores for each sentence.

Annotators assessed adaptation errors only when no meaning transfer errors (PRN, TRM, GSMIS) were present, as style or tone evaluation is irrelevant when meaning is lost. To prevent over-penalizing systems that preserved meaning but erred in adaptation, adaptation scores were weighted at 50\% in the segment total error score (\texttt{SEGS}) calculation.

Three physical copies of the Excel sheet were printed (one for each system) as annotators preferred working on paper. Each contained the same DA-MSA pairs, differing only in the translation engine output. Annotators marked scores manually, and data were later transferred to the digital Excel template.

\section{Results Analysis}
\subsection{Inter Annotator Agreement}
Inter-Annotator Agreement (IAA) measures consistency between evaluators, with high scores indicating a reliable and reproducible annotation framework. 
In our study, we employed Quadratic Weighted Cohen’s Kappa (QWK) because it effectively handles ordinal severity ratings by assigning greater penalties to larger disagreements, whereas, the ordinary Cohen’s κ treats all disagreements equally, unweighted κ, e.g. 0 vs 1 = 0 vs 2. In addition, quadratic weights are standard for ordinal ratings
\cite{cohen1968weighted}.
This makes it particularly suitable for error severity judgment on our 0-2 scale, where the magnitude of disagreement matters. 
%both count the same
The IAA scores are presented in Table~\ref{tab:cohens_kappa}.

As shown in Figure~\ref{fig:annotation-tree}, Meaning Transfer comprises three fine-grained error types (PRN, TRM, and GSMIS). Accordingly, for Meaning Transfer IAA, we compute inter-annotator agreement on an aggregated severity score obtained by summing PRN, TRM, and GSMIS. This reflects annotator agreement on overall meaning-transfer impact rather than on individual sub-error categories. Similarly, SEGS represents the segment-level total error score (produced by summing all five error types’ scores), and the corresponding IAA reflects agreement on overall segment-level error severity.

The IAA scores varied meaningfully across error categories:
Fluency showed strong agreement for most MT systems, likely due annotators’ Arabic language expertise, which minimized inaccurate judgments. 
Meaning Transfer achieved robust agreement, likely due to the clear criteria regarding preservation of original meaning, as established in the Ara-HOPE taxonomy decision tree (Figure~\ref{fig:annotation-tree}). Those definitions were particularly refined before the full evaluation during our pilot testing and feedback sessions. This consistency indicates a shared understanding of meaning preservation requirements among our annotators. 
Adaptation yielded the lowest agreement (0.122-0.28) due to its inherently subjective nature. 

A translation has an adaptation error when it seems unnatural or inappropriate to the context, situation, or intent of the source. Unlike fluency or meaning-transfer errors, which can often be localized to specific parts of the translation, adaptation errors require more holistic, and usually subjective judgment. One example is the dialectal sentence: \\

\foreignlanguage{arabic}{عم بحكي من واقع تجربة كانت رح تقضي ع حلم حياتي بس الحمد لله الحب اقوى} ("I am speaking from experience. It almost destroyed my life's dream, but thank God, love is stronger"), which the Jais model translated as\\

\foreignlanguage{arabic}{أنا أتحدث بناءً على تجربتي الشخصية التي كادت تدمر حلمي الأكبر، ولكن بحمد الله، أثبت الحب أنه أقوى بكثير.} (”I speak from my personal experience, which almost destroyed my biggest dream, but thank God, love proved to be much stronger.”)\\

While both annotators agreed that the translation is fluent and preserves the source meaning, one judged it to be slightly less natural in MSA (ADP = 1, minor adaptation error), whereas the other annotator found it acceptable and contextually appropriate.

The annotator disagreement likely stems from the rendering of phrases suchas \foreignlanguage{arabic}{بس الحمد لله الحب اقوى} (“But thank God, love is stronger”) which was translated as \foreignlanguage{arabic}{ولكن بحمد الله، أثبت الحب أنه أقوى بكثير}. (”But thank God, love proved to be much stronger”), which may be seen as stylistically over-emphatic or slightly less natural in MSA for this context.

\begin{table}[htbp]
\centering
\small 
\begin{tabular}{lccc}
\hline
Error Type  & Jias & GPT3.5 & NLLB200 \\
\hline
Fluency   & 0.507 & 0.552 & 0.368 \\
Meaning Transfer   & 0.529 & 0.629 & 0.554 \\
Adaptation   & 0.171 & 0.122 & 0.280 \\
\hline
SEGS & 0.608 & 0.629 & 0.500 \\
\hline\hline
\end{tabular}
\caption[]
{Quadratic Weighted Kappa (QWK) scores across models and error types. Meaning Transfer IAA is computed on aggregated PRN+TRM+GSMIS severity; SEGS reflects total segment-level error severity.}
\label{tab:cohens_kappa}
\end{table}

The IAA scores for the three systems in our study on the segment-level total error score (SEGS) was 0.5 to 0.629, indicating reasonably consistent human evaluation. 

Prior work, including \cite{8d20e0b8-89d8-3d65-bcf5-8c19d56ec4ab}, considers kappa scores in the range of 0.41-0.60 to indicate a moderate level of agreement and 0.61-0.80 to indicate substantial agreement. Nevertheless, it is important to note that the interpretation of standard Kappa and Quadratic Weighted Kappa varies considerably.

Aggregating fine-grained error categories into a composite severity score reduces sparsity and stabilizes the marginal distributions, which is known to improve reliability of composite measures \cite{fleiss2003statistical}. % (Fleiss et al., 2003). 
Moreover, Quadratic Weighted Kappa assigns smaller penalties to near disagreements \cite{cohen1968weighted}, so cross-category disagreements often translate into small ordinal differences after aggregation. Consequently, agreement computed on aggregated segment-level scores can exceed that of individual categories.
%, consistent with MQM- and post-editing-based evaluation practices (Lommel et al., 2013; Gladkoff and Han, 2022).

\begin{figure*}[t]
\begin{center}
\includegraphics[width=0.95\textwidth]{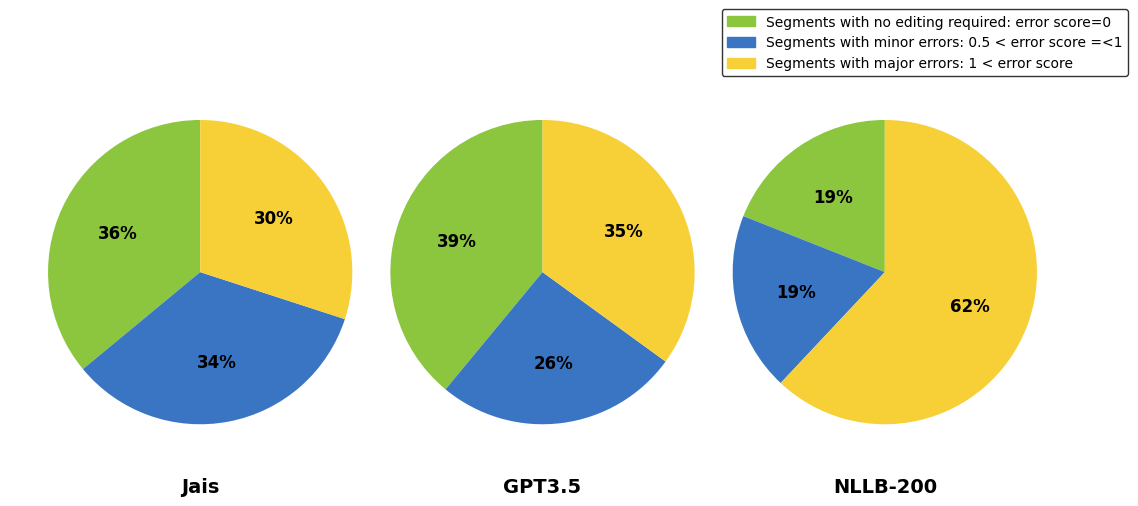} % Rotate by 90 degrees and scale to full page width
\end{center}
\caption[Error Severity Across Translation Systems]
{Comparison of error severity distribution among Jais, GPT-3.5, and NLLB-200, highlighting proportions of major, minor, and no-error translations.}
\label{fig:error-severity}
\end{figure*}

\subsection{Quantitative Error Analysis}
The quantitative analysis below presents a comparison of MT system performance, highlighting differences across key evaluation criteria. This assessment identifies each system’s strengths and weaknesses, as well as both unique and shared challenges across systems.

\textbf{Error Severity Analysis}
This analysis examines translation error severity for 205 sentences produced by the three MT systems using the accumulated sentence total error scores (SEGS). It is essential for understanding how each model handles DA-MSA translation and for identifying systems that generate higher-quality outputs that require minimal post-editing.

SEGS is calculated as the sum of the scores assigned to each of the five error types per sentence, yielding a range of 0-4 per sentence. For comparison, SEGS values were grouped as follows: segments requiring no editing (SEGS = 0), segments with minor errors ($0.5 < \text{SEGS} \leq 1$), and segments with major errors ($\text{SEGS} > 1$).

Figure~\ref{fig:error-severity} shows clear differences among the systems. For Jais, 36\% of segments required no editing, 34\% had minor errors, and 30\% had major errors. GPT-3.5 performed slightly better, with 39\% of segments requiring no editing, fewer minor errors (26\%), but slightly more major errors (35\%). NLLB-200 performed the worst, with only 19\% of segments requiring no editing and 62\% containing major errors.

These results reflect differences in model architecture and training. Jais, as an Arabic-centric model, handles DA-specific nuances well but still struggles with complex segments. GPT-3.5’s general-purpose design provides balanced performance but lacks Arabic specialization. The strong performance of Jais and GPT-3.5 underscores the relative strength of decoder-only models over encoder-decoder models like NLLB-200.

\begin{figure*}[t]
\begin{center}
\includegraphics[width=\textwidth]{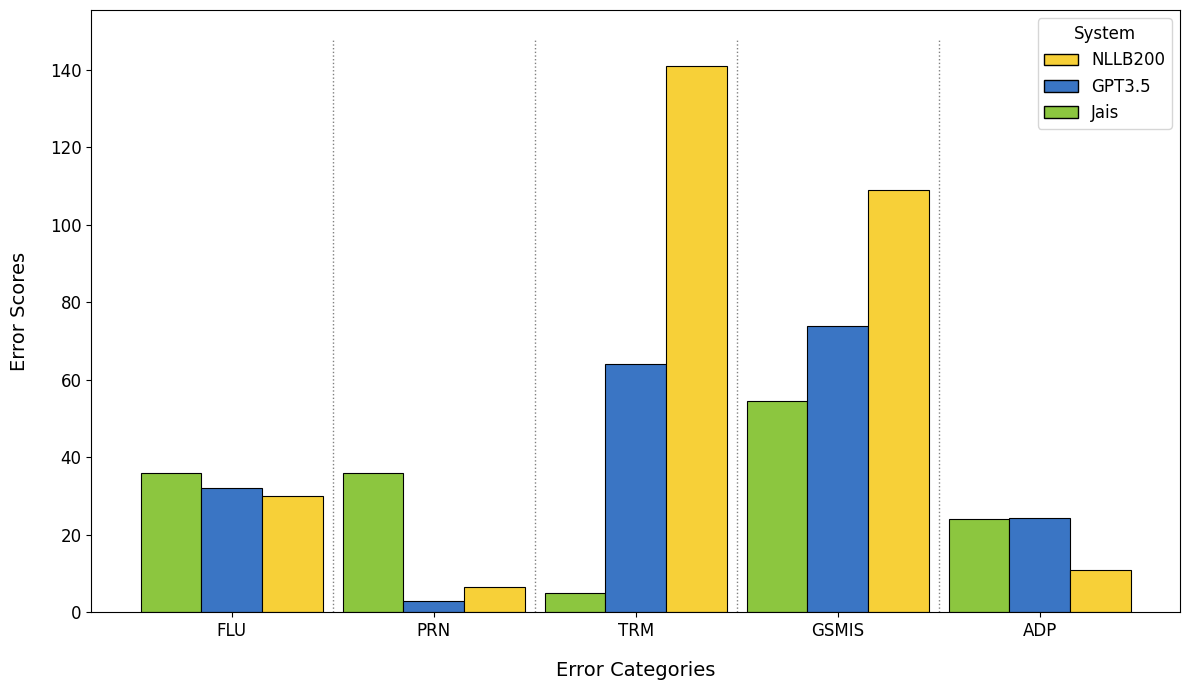} % Rotate by 90 degrees and scale to full page width
\end{center}
\caption[Error Scores by Category and Translation System]
{Comparison of Jais, GPT-3.5, and NLLB-200 models’ error scores across error categories}
\label{fig:error-pattern}
\end{figure*}

\textbf{Error Pattern}
Examining the error distributions in Figure~\ref{fig:error-pattern} provides key insights into each system’s approach to DA-MSA translation challenges. NLLB-200 exhibits a high frequency of dialect-specific term (TRM) and semantic preservation (GSMIS) errors, highlighting its encoder–decoder architecture’s difficulty with idiomatic expressions, cultural references, and context-dependent meanings specific to Levantine Arabic. These issues stem from limited domain-specific tuning and insufficient exposure to regional culturally embedded expressions, which are essential for accurate meaning transfer.

Jais, despite its Arabic-specialized pretraining, struggles with proper names (PRN). This might be because the system tends to over-normalize or excessively standardize names transliterated in social media DA content, leading to the loss of original or intended forms. However, its strong performance with dialect-specific terminology reflects effective handling of DA morphological variations, likely due to its extensive pretraining on Arabic data.

GPT-3.5 shows a balanced error distribution across many categories, suggesting that its large multilingual pretraining and sheer parameter size help offset the lack of explicit Arabic dialectal training. This is particularly evident in challenging cases involving code-mixing (where words from two languages or dialects are used in the same sentence) and pragmatic shifts, which require interpreting meaning based on context, tone, or speaker intent rather than literal words. These cases require advanced comprehension, and the model performs well in them even though it was not specifically trained on Arabic dialects.

The consistently low fluency (FLU) error rates across systems indicate that syntactic reconstruction from DA to MSA is less challenging than lexical-semantic transfer\footnote{Lexical-semantic transfer refers to the mapping of words and their intended meaning from one dialect or language variety to another.}. Low adaptation (ADP) error rates across all systems are largely due to this error type being assessed only when meaning transfer errors (PRN, TRM, GSMIS) are absent. When a text fails to convey its intended meaning, evaluating its stylistic or cultural appropriateness becomes less relevant.

\begin{figure*}[t]
\begin{center}
\includegraphics[scale=1.6, width=\textwidth]{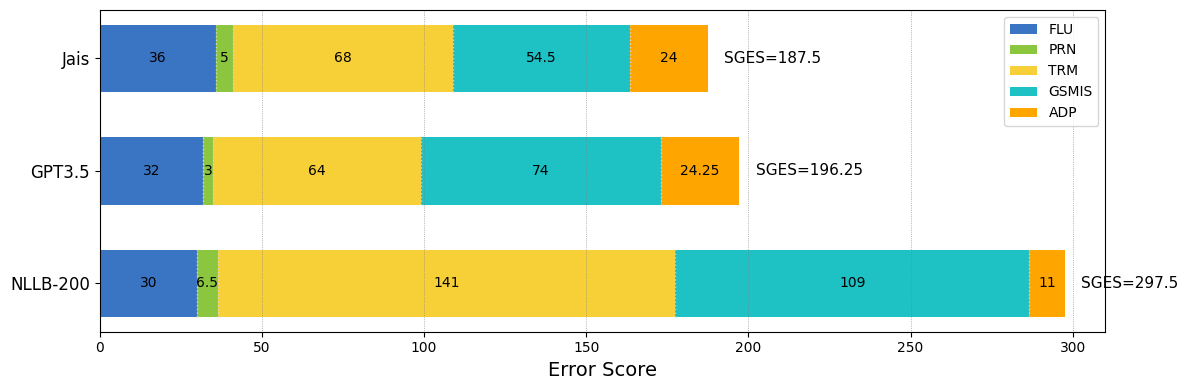} % Rotate by 90 degrees and scale to full page width
\end{center}
\caption[Detailed Error Distribution by Category and System]
{Visualization of accumulated error scores across fluency, meaning transfer, and adaptation error categories for Jais, GPT-3.5, and NLLB-200 in DA-MSA translation.}
\label{fig:error-dist}
\end{figure*}

A complementary view is provided in Figure~\ref{fig:error-dist}, which shows the exact error distribution for each system. The total error scores for all sentences (SEGS). It is clear that NLLB-200 has a significantly higher total error score (297.50) compared to GPT-3.5 (196.25) and Jais (187.50), with Jais producing the least meaning transfer errors. Notably, TRM and GSMIS errors constitute the majority of total errors across all systems, indicating that future MT development for DA-MSA should prioritize improvements in semantic accuracy and dialect-specific terminology handling.

\subsection{Qualitative Error Analysis}

% While the previous section presented quantitative results to highlight statistical differences in system performance, 
This qualitative analysis offers concrete examples of two common error types to illustrate how translation quality varies across MT systems, focusing on challenging linguistic phenomena.
% Although the project emphasizes computational linguistics over pure linguistic analysis, comparative qualitative evaluation provides actionable insights to guide future DA-MSA machine translation development.

\subsubsection{Analysis of Dialectal Terminology Errors}
Addressing dialect-specific terms (TRM) is a core challenge in DA-MSA translation due to vocabulary mismatches between informal dialects and formal Arabic. Table~\ref{tab:error-trm} presents a sentence containing a unique dialect-specific term. The word \foreignlanguage{arabic}{بيجنن} ("amazing" in Syrian Arabic) lacks a direct MSA equivalent, requiring systems to interpret its meaning contextually instead of relying on word-for-word mapping. Unlike NLLB-200, Jais, with its Arabic-centric pretraining, was better at recognizing this regional idioms. GPT-3.5, though not DA-specialized, leverages its broad language understanding to connect dialect terms to contextually appropriate MSA words (translating \foreignlanguage{arabic}{بيجنن} as \foreignlanguage{arabic}{ممتاز} "excellent"). These findings demonstrate that success with TRM errors depends heavily on exposure to diverse, dialect-rich data that treats dialectal phrases as meaningful units to be interpreted contextually, rather than treating them as isolated words.

\begin{table}[t]
\centering
\setlength{\tabcolsep}{4pt}
\renewcommand{\arraystretch}{1.2}

\begin{tabular}{|p{0.95\linewidth}|}
\hline
\textbf{Source DA: }\foreignlanguage{arabic}{اليوم الجو يجنن، لازم نطلع نتمشى} \\
\textbf{Gold MSA: } \foreignlanguage{arabic}{اليوم الطقس رائع، يجب أن نخرج للنزهة} \\
``The weather is wonderful today; we should go out for a walk.'' \\
\hline
\textbf{Jais: } \foreignlanguage{arabic}{اليوم الطقس رائع، يجب أن نذهب للنزهة}\\
``The weather is wonderful today; we should go for a walk.'' \\
\hline
\textbf{GPT-3.5: }\foreignlanguage{arabic}{اليوم الطقس ممتاز، يجب أن نخرج للنزهة} \\
``The weather is excellent today; we should go out for a walk.'' \\
\hline
\textbf{NLLB-200: }\foreignlanguage{arabic}{اليوم الطقس مجنون، يجب أن نخرج} \\
``Today's weather is crazy; we should go out for a walk.'' \\
\hline
\end{tabular}
\caption[Analysis of Dialect-Specific Terms (TRM) Errors]
{Evaluation of how models handle dialect-specific terms (TRM), highlighting translation challenges in translating informal expressions to MSA.}\label{tab:error-trm}

\end{table}

\subsubsection{Analysis of Adaptation Errors}
Table~\ref{tab:error-gsmis} presents an example illustrating how the different systems preserved the intent and tone of the source when translating to MSA. ADP errors highlight the gap between DA’s context-dependent expressions and MSA’s formality. For instance, translating \foreignlanguage{arabic}{شو قصتك} (casual ``What’s wrong?'') requires capturing the speaker’s intent rather than just the literal words. NLLB-200 translates this as \foreignlanguage{arabic}{ما هي قصتك} (``What is your story?''), focusing on a literal interpretation. Jais, benefiting from exposure to both dialect and MSA data, uses the conventional MSA phrase \foreignlanguage{arabic}{ما الخطب} (``What’s the matter?''). GPT-3.5 chooses the phrasing \foreignlanguage{arabic}{ماذا حدث} (``What happened?''). These differences show that reducing ADP errors requires systems to prioritize intent and context over lexical mapping, treating DA as a distinct communication style with its own contextual rules rather than simply a variation of MSA.

\begin{table}[t]
\centering
\setlength{\tabcolsep}{4pt}
\renewcommand{\arraystretch}{1.2}

\begin{tabular}{|p{0.95\linewidth}|}
\hline
\textbf{Source DA: }\foreignlanguage{arabic}{شو قصتك ليش معصب} \\
\textbf{Gold MSA: } \foreignlanguage{arabic}{ما الأمر؟ لماذا تبدو غاضبًا؟} \\
``What's wrong? Why do you look angry?'' \\
\hline
\textbf{Jais: } \foreignlanguage{arabic}{ما الخطب؟ لماذا تبدو غاضبًا}\\
``What's the matter? Why do you look angry?'' \\
\hline
\textbf{GPT-3.5: }\foreignlanguage{arabic}{ماذا حدث؟ لماذا تبدو غاضبًا؟} \\
``What happened? Why do you look angry?'' \\
\hline
\textbf{NLLB-200: }\foreignlanguage{arabic}{ما هي قصتك؟ لماذا أنت غاضب؟} \\
``What is your story? Why are you angry?'' \\
\hline
\end{tabular}
\caption[Analysis of Adaptation (ADP) Errors]
{Comparison of model performance on Adaptation (ADP) errors.}\label{tab:error-gsmis}

\end{table}

Overall, our analyses emphasize that effective DA-MSA translation depends on training strategies that prioritize (1) comprehension of dialect phrases and (2) preservation of speaker intent across registers. The qualitative analysis above shows that low-resource MT is a creative task, involving sub-tasks like sentiment analysis and formality adaptation, which go beyond simple lexical mapping.

\section{Conclusion}
This paper introduces the \textit{Ara-HOPE} framework as a human-centric approach for evaluating DA-MSA translation, successfully fulfilling its intended objectives through a specialized error taxonomy, an efficient annotation workflow, and a comparative evaluation of different MT systems. The \textit{five-category error classification} system effectively captures translation challenges unique to DA, while the \textit{decision tree protocol} improves annotation consistency. Quantitative findings reveal significant differences in performance among Arabic-centric (Jais), general-purpose (GPT-3.5), and baseline (NLLB-200) systems, with dialect-specific terminology and semantic preservation identified as key challenges. By systematically addressing the complexities of DA-MSA translation assessment through rigorous human evaluation, Ara-HOPE establishes reproducible standards for Arabic MT assessment and provides actionable insights to guide future MT systems development \footnote{This work is in line with our DA-MSA MT work at \cite{alabdullah2025advancingdialectalarabicmodern} where we examined LLM prompting vs finetuning for Levantine, Egyptian, and Gulf dialects to MSA translation.}.

% \subsection{Abdul to relocate:}
% \[
% \text{MT}_s = \text{PRN}_s + \text{TRM}_s + \text{GSMIS}_s
% \]

% \small 
% \[
% \text{SEGS}_s = \text{FLU}_s + \text{PRN}_s + \text{TRM}_s + \text{GSMIS}_s + 0.5 \cdot \text{ADP}_s
% \]

% \[
% \kappa_{\text{QWK}}\big(\text{MT}^{(1)}, \text{MT}^{(2)}\big), \quad
% \kappa_{\text{QWK}}\big(\text{SEGS}^{(1)}, \text{SEGS}^{(2)}\big)
% \]

% Quadratic weighted kappa:
% \[
% w_{ij} = 1 - \frac{(i - j)^2}{(K - 1)^2},
% \]
% where $i$ and $j$ are the ordinal ratings assigned by the two annotators and $K$ is the number of possible rating levels.

\section{Limitations}
In this work, we only used zero-shot prompting to generate translations. Future research could explore human evaluation of translations produced using alternative prompting strategies, such as few-shot or chain-of-thought prompting. Additionally, the human annotation process was time-consuming. Future work could consider using LLM-as-a-judge approaches to partially or fully automate the annotation process. Our work focused on human evaluation, and we did not investigate the correlation between human judgment and automatic evaluation metrics, lexicon-based and neural-embedding based, like BLEU and BERTScore. We leave that for future work.

\section*{Acknowledgment}
We would like to thank our volunteer human annotators, Heba Sado Jazea and Rania Mohammad Daher, for their invaluable time and effort throughout the annotation process. Their careful annotations and insightful feedback were essential to establishing our Ara-HOPE framework. We also thank the VarDial reviewers for the valuable comments.

\bibliography{custom}
%\bibliography{latex/refs}

\appendix

\begin{figure*}
\centering
\includegraphics[angle=270, width=0.9\textwidth]{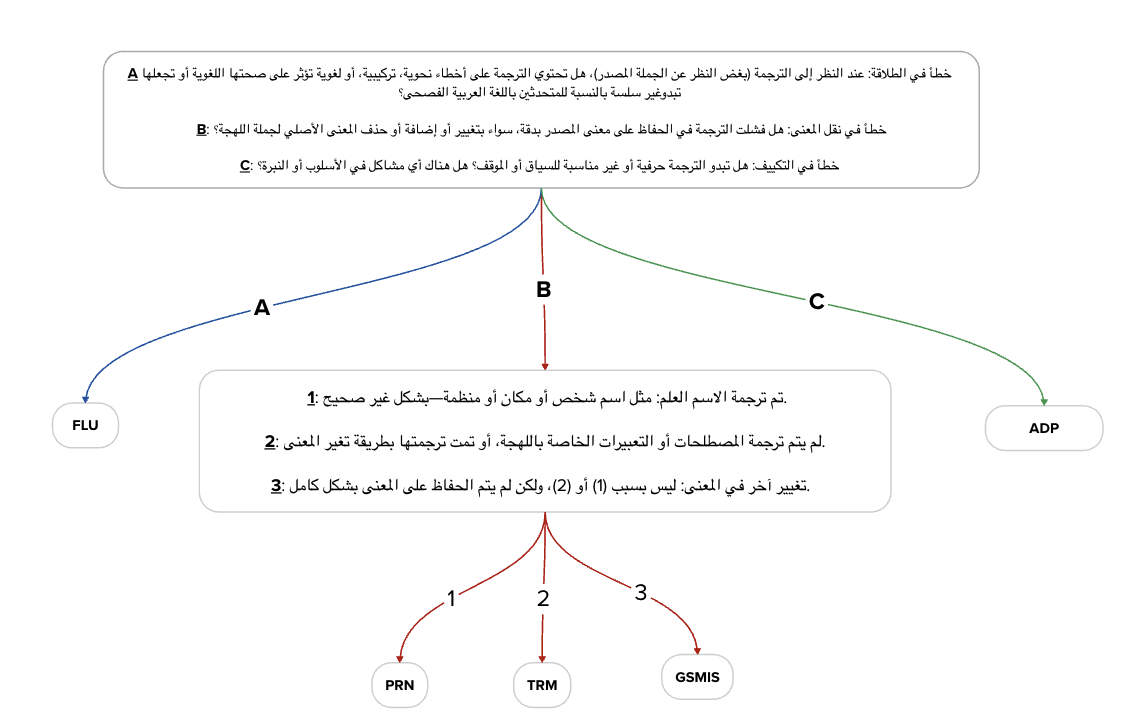} % Only width
\caption[Ara-HOPE Annotation Decision Tree (Arabic)]
{The Arabic version of the Ara-HOPE Annotation Decision Tree. A structured decision tree guiding annotators through error classification for evaluating DA-MSA translations using the Ara-HOPE framework.}
\label{fig:arabic-annotation-tree}
\end{figure*}

\begin{figure*}
\centering
\includegraphics[angle=270, width=0.95\textwidth]{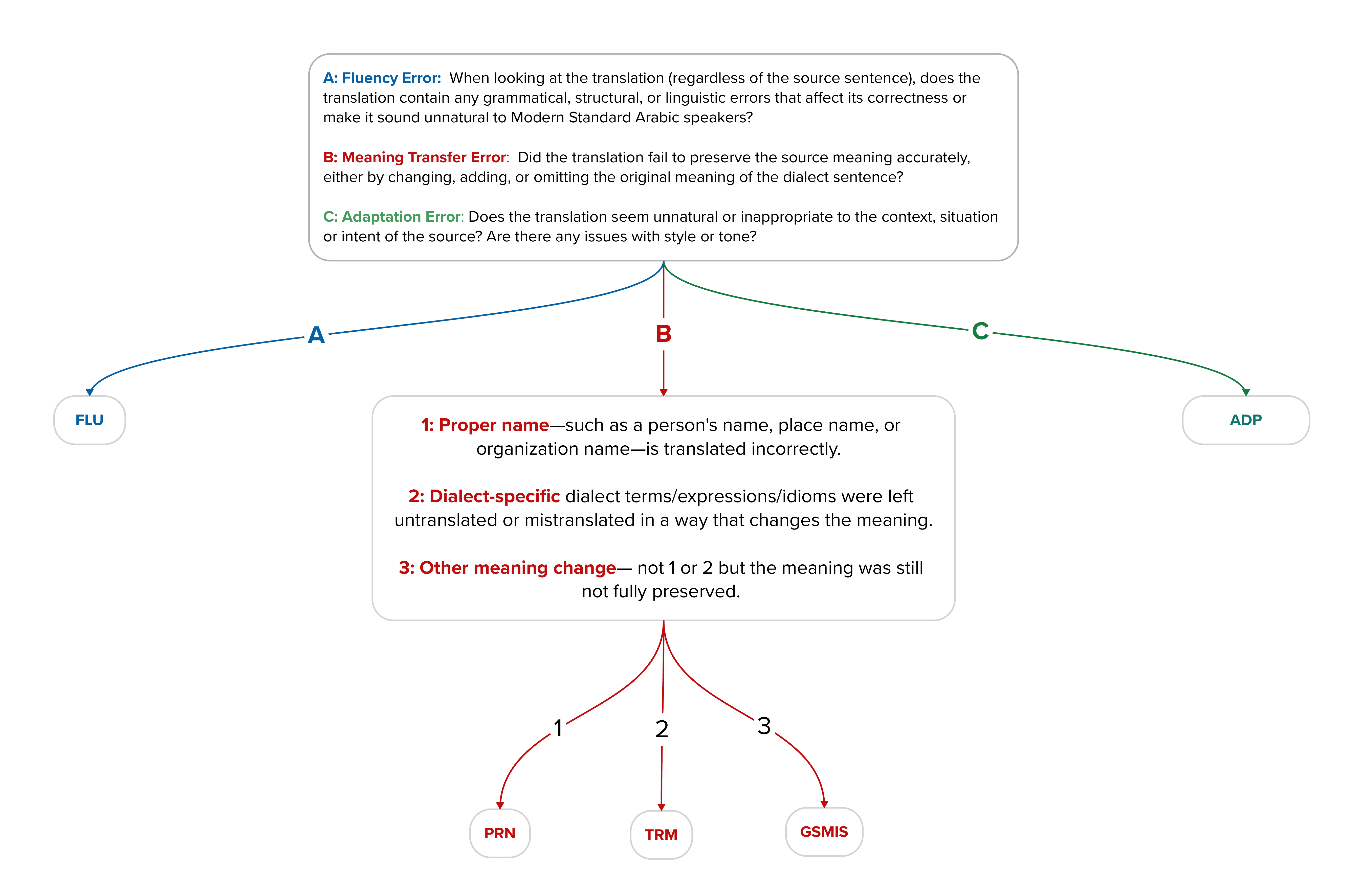} % Only width
\caption[Ara-HOPE Annotation Decision Tree]
{Structured decision tree guiding annotators through error classification for evaluating DA-MSA translations using the Ara-HOPE framework.}
\label{fig:annotation-tree}
\end{figure*}

\begin{figure*}
\centering
\includegraphics[width=\textwidth]{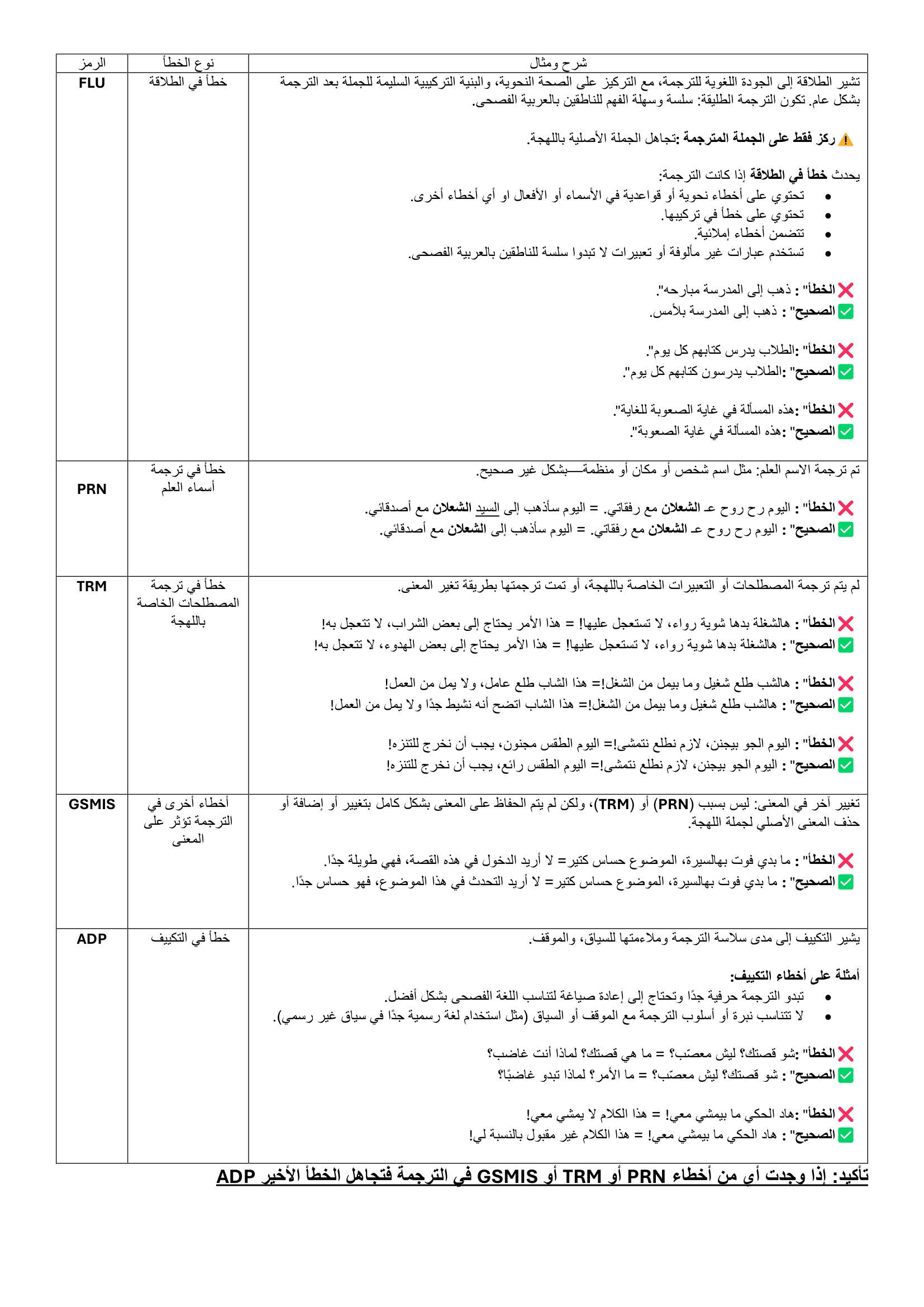} % Only width
\caption[Ara-HOPE Annotation Guidelines]
{The annotation guidelines provided to human annotators explain each error type with illustrative examples, assisting them in using the Ara-HOPE Annotation Decision Tree.}
\label{fig:annotation-guidelines}
\end{figure*}

% \subsection{Parameter-Efficient Fine-tuning for Low-Resource Languages}

% Recent advances in parameter-efficient fine-tuning have revolutionized the adaptation of large language models (LLMs) for low-resource machine translation. Low-Rank Adaptation (LoRA) has emerged as a dominant approach, enabling efficient fine-tuning by updating only a small subset of model parameters. Cao et al. (2024) introduce a language-specific LoRA fine-tuning strategy that models intrinsic language-dependent subspaces. Their experiments show that these targeted LoRA modules significantly improve multilingual NMT performance, with the largest gains observed in low-resource language pairs \cite{cao2024}. Similarly, Language-Specific Fine-Tuning with Low-rank adaptation (LSFTL) has shown promising results by selectively optimizing the multi-head attention and feed-forward networks in Transformer layers while preserving most original parameters.

% Beyond single-stage adaptation, researchers have explored more sophisticated approaches. Gao et al. (2024) proposed a novel two-step fine-tuning framework: first adapt the parent model using child-language monolingual data, then transfer to the low-resource target pair via fine-tuning. They reported significant improvements on multiple low-resource translation tasks \cite{gao-etal-2024-novel}. Quantization techniques further reduce computational requirements, with Ibrahim (2024) successfully implementing 8-bit quantization alongside LoRA to fine-tune LLaMA-3 for translating between Arabic dialects and MSA \cite{ibrahim2024}.

\end{document}